\documentclass[runningheads]{llncs}

\PassOptionsToPackage{table}{xcolor}
 
\usepackage[mobile]{eccv}

\usepackage{eccvabbrv}

\usepackage{graphicx}
\usepackage{booktabs}
\usepackage{amsmath}
\usepackage{amssymb}
\usepackage{multirow}
\usepackage{tabularx}
\usepackage{xstring}
\usepackage{pifont}
\usepackage{colortbl}
\usepackage{subcaption}
\usepackage{algorithm}
\usepackage{algpseudocode}
\usepackage{newfloat}
\usepackage{listings}
\usepackage{tikz}
\usetikzlibrary{matrix, positioning, arrows.meta}
\usetikzlibrary{calc}

\DeclareCaptionStyle{ruled}{labelfont=normalfont,labelsep=colon,strut=off} 
\floatstyle{ruled}
\newfloat{listing}{tb}{list}{}
\floatname{listing}{Listing}

\usepackage[accsupp]{axessibility}  

\newcommand{\bdelta}{\boldsymbol\delta}
\newcommand{\bx}{\mathbf{x}}
\newcommand{\bmu}{\mathbf{\boldsymbol\mu}}

\newcommand{\btheta}{\boldsymbol\theta}

\newcommand{\bB}{\mathbf{B}}
\newcommand{\bA}{\mathbf{A}}

\definecolor{myred}{RGB}{128,128,128}
\definecolor{mygreen}{RGB}{0,0,0}

\newcommand{\ours}{\textbf{\textsc{ES-VP}}\,}

\usepackage{xr-hyper}
\usepackage{hyperref}

\usepackage{orcidlink}

\begin{document}

\title{\ours: Energy-Shaped Dynamic Visual Prompting for Efficient Model Adaptation} 

\titlerunning{ES-VP}


\author{%
Can Jin\inst{1}$^{*}$ \and
Ying Li\inst{2}$^{*}$ \and
Jingchen Sun\inst{3} \and
Hongwu Peng\inst{4} \and
Jiahui Zhao\inst{5} \and
Yang Zhou\inst{1} \and
Lei Li\inst{6} \and
Dimitris N. Metaxas\inst{1}}

\institute{%
Rutgers University \and
Zhejiang University \and
University at Buffalo, SUNY \and
Adobe Research \and
University of Connecticut \and
Washington University}

\authorrunning{C. Jin et al.}



\maketitle
{\let\thefootnote\relax\footnotetext{$^{*}$ Equal contribution.}}

\begin{abstract}
Visual prompting (VP) has emerged as a parameter-efficient method for adapting pre-trained models to downstream tasks. However, existing approaches encounter a trade-off between flexibility and efficiency. Some methods apply a fixed prompt to all images, ignoring individual image characteristics, while others introduce auxiliary networks to generate diverse prompts. Although the latter can improve performance, it also significantly increases parameter usage and the potential for overfitting to specific datasets. Furthermore, the auxiliary networks, combined with inherent biases in pre-trained models, limit scalability and generalization. In this paper, we propose Energy-Shaped Visual Prompting (\ours), a novel approach that generates image-specific prompts using low-rank initialization and energy-guided dynamic adaptation, achieving superior performance with fewer parameters compared to single-prompt methods. \ours directly utilizes the pre-trained model for adaptive prompt generation, ensuring both parameter efficiency and improved generalization. Extensive experiments conducted on five architectures across fifteen datasets demonstrate that \ours consis-tently outperforms current state-of-the-art (SOTA) single and diverse VP methods. For instance, using the CLIP architecture across four datasets, \ours outperforms the SOTA method DAM-VP by an average of 2.6\% in accuracy while utilizing 590$\times$ fewer VP parameters, thereby establishing a new benchmark for efficient and generalizable model adaptation.

\end{abstract}
\vspace{-4mm}
\section{Introduction}\label{section_introduction}
\vspace{-2mm}
A wide range of computer vision (CV) and natural language processing (NLP) tasks leverage the flexibility and power of large-scale pre-trained models by adapting these foundational models to suit diverse task-specific applications \cite{liu2021swin, dosovitskiy2020image, ridnik1imagenet, NEURIPS2020_1457c0d6,li2026multiple}. Techniques like in-context learning \cite{NEURIPS2020_1457c0d6, shin-etal-2020-autoprompt,liu2022few,liu2026bcl} and prompt tuning strategies \cite{liu2023pre} leverage well-designed input templates to improve model performance across diverse tasks. This paradigm has inspired visual prompting (VP) within computer vision, where it is employed to adapt pre-trained vision models by modifying inputs at the pixel level or through output transformations, rather than altering the model's internal weights or structure \cite{bahng2022exploring, chen2023understanding, tsao2024autovp,huang2023diversity,caisample}. Unlike conventional transfer-learning methods that require full fine-tuning, linear probing (LP) of trainable layers, or zero-shot learning \cite{radford2021learning}, VP retains the model's pre-trained backbone by adapting it to new tasks with minimal parameter tuning.

\begin{figure}[t]
    \vspace{-2mm}
    \centering
    \includegraphics[width=0.9\linewidth]{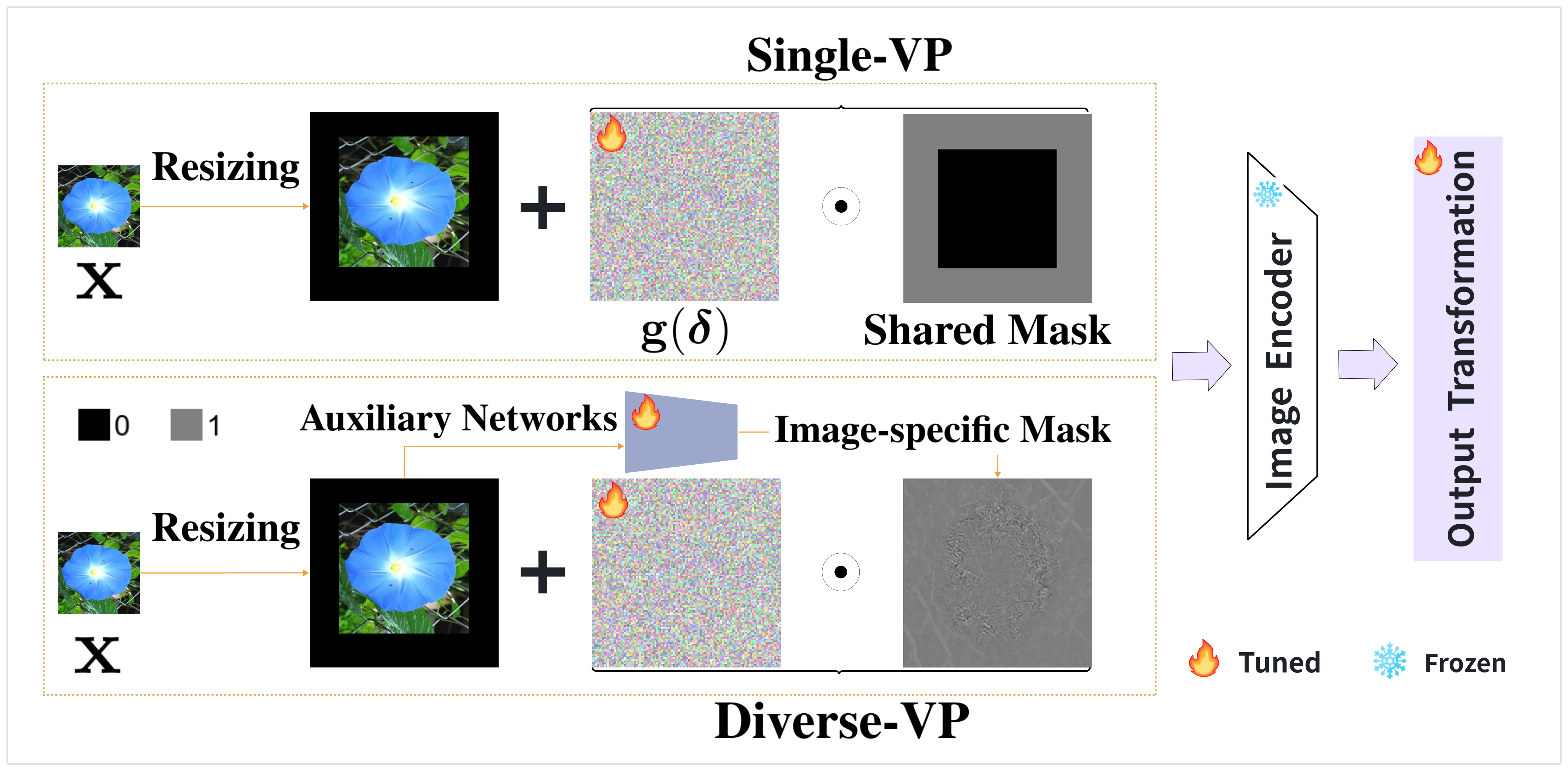}
    \caption{\textbf{Existing VP methods.} Existing VP methods can be categorized into two groups: Single-VP and Diverse-VP methods. $\boldsymbol{g}(\boldsymbol{\delta})$ is the VP parameters.}
    \label{figure_existing_method}
    \vspace{-5mm}
\end{figure}

As shown in Figure \ref{figure_existing_method}, existing VP methods can be broadly categorized into two main strategies: \ding{182} Employing a \textbf{single universal visual prompt} (Single-VP) for all images in the downstream task \cite{bahng2022exploring,chen2023understanding,tsao2024autovp,jin2025visual,jin2025lorvp}. Although this method is parameter-efficient, it fails to account for the considerable variability among images, resulting in prompts that lack the specificity necessary to capture the unique characteristics of individual images. Consequently, this approach limits the discriminative capability of deep models. \ding{183} Using \textbf{diverse visual prompts} (Diverse-VP) for different images, typically involving multiple VP initializations or auxiliary networks to generate image-specific prompts \cite{huang2023diversity,caisample}. While this approach addresses the limitations of single-prompt methods by enhancing specificity, the additional parameters introduced by multiple prompts or auxiliary networks conflict with the objective of parameter-efficient adaptation and can substantially hinder the scalability of pre-trained models. Thus, we ask: \textit{Is there an effective visual prompting method capable of generating image-specific prompts while maintaining parameter efficiency?}

\begin{figure}[t]
    \vspace{-1mm}
    \centering
    \includegraphics[width=0.98\linewidth]{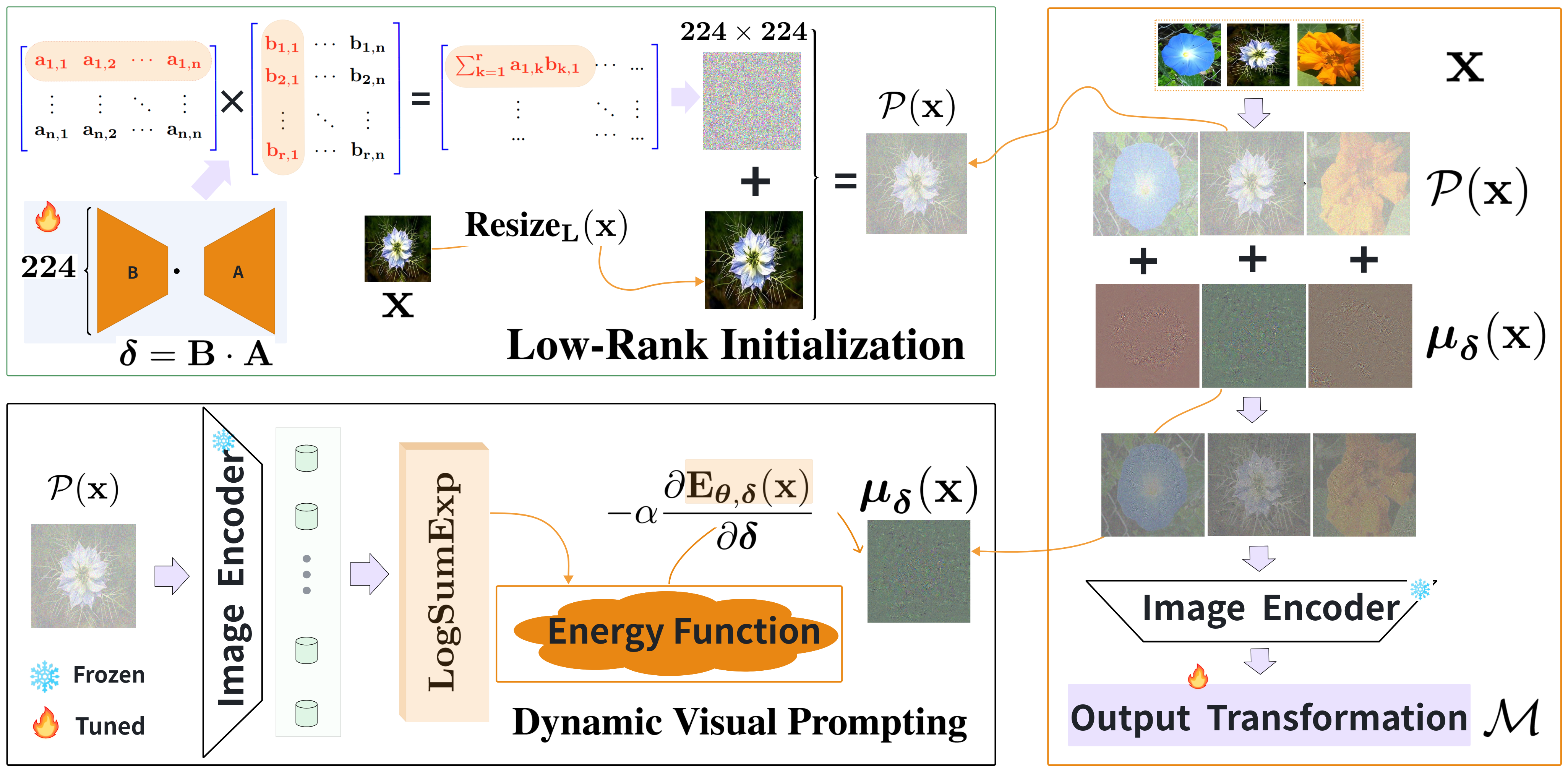}
    \caption{\textbf{Overview of \ours.} We establish a single universal VP via a low-rank prior $\bdelta$ to capture global information, and derive an image-specific dynamic adaptation $\boldsymbol{\mu}_{\boldsymbol{\delta}}(\mathbf{x})$ using an energy function to form the final prompted image $\mathcal{P}(\mathbf{x})$. Both the low-rank prior and the dynamic adaptation are jointly optimized via meta-learning.}
    \label{figure_our_method}
        \vspace{-4mm}
\end{figure}

In this paper, we introduce adaptive Energy-Shaped Visual Prompting (\ours), a method designed to generate image-specific visual prompts (VPs) without increasing the number of VP parameters or requiring auxiliary networks. As illustrated in Figure \ref{figure_our_method}, \ours comprises two primary components: \ding{182} A universal low-rank prompt initialization, which is applied uniformly to all images, capturing global task characteristics and common dataset patterns. To enhance parameter efficiency, this initialization utilizes two low-rank matrices inspired by proven low-rank adaptation techniques and the intrinsic low-rank structure observed in deep neural networks \cite{hu2021lora, aghajanyan2021intrinsic, allen2019convergence}. \ding{183} An image-specific, parameter-free dynamic prompt adjustment derived from the initial prompt and tailored to each individual image. This dynamic adjustment captures unique and discriminative image features, enabling detailed adaptability. We generate these adjustments using a label-independent energy function from the pre-trained model, allowing automatic and adaptive VP generation for each image without additional networks, thus reducing the risk of overfitting. The low-rank prior and dynamic adaptation are jointly optimized via meta-learning to ensure rapid convergence and highly effective prompt generation. Furthermore, the energy-based approach provides flexibility and adaptability compatible with any pre-trained classifier, effectively shaping the energy landscape for out-of-distribution samples and improving VP generalization \cite{grathwohl2019your, liu2020energy}.

We conduct extensive experiments using five different architectures to evaluate the effectiveness of \ours across various model designs. Experiments performed on fifteen datasets consistently demonstrate that \ours surpasses current SOTA Single-VP and Diverse-VP methods. For example, with the CLIP architecture \cite{radford2021learning}, \ours achieves an accuracy improvement of \(2.6\%\) compared to the SOTA diverse-VP method DAM-VP \cite{huang2023diversity} on four datasets, while utilizing $\times$ 590 fewer VP parameters. Additionally, \ours achieves \(4.5\%\) higher accuracy than the Single-VP method AutoVP \cite{tsao2024autovp} on ten datasets using ResNet-18 \cite{he2016deep}, with $\times$ 12 fewer VP parameters. In summary, our contributions are:
\begin{itemize}
    \item We introduce a novel adaptive energy-shaped VP method capable of generating discriminative, image-specific visual prompts with significantly fewer parameters compared to existing Single and Diverse VP methods.

    \item Extensive experiments across five architectures and fifteen datasets consistently demonstrate the superior in-distribution and out-of-distribution generalization capabilities of \ours to SOTA Diverse-VP methods, highlighting its potential effectiveness for real-world applications.

    \item Comprehensive analyses and ablation studies provide insights into the effectiveness and efficiency of \ours, offering valuable guidance for using energy-based methods in parameter-efficient dynamic adaptation.
\end{itemize}

\section{Related Works}\label{section_related_works}
\subsection{Visual Prompting}
Prompting, initially developed for NLP, is a technique that adapts pre-trained models to specific tasks by modifying their inputs \cite{shin-etal-2020-autoprompt, liu2022few, liu2023pre,jin2025visual,jin2025lorvp}. This method was subsequently extended to CV through VP, introduced by \cite{bahng2022exploring}, where adjustable parameters are directly embedded into visual inputs to guide pre-trained models. A standard VP framework consists of two main components: input design, where prompts are integrated into the input image, and output transformation, which modifies model outputs to suit new tasks \cite{bahng2022exploring, tsai2020reprogramming, tsao2024autovp, caisample}. Several VP strategies have been proposed, either applying a single visual prompt (Single-VP) universally to the downstream dataset or utilizing diverse visual prompts (Diverse-VP) customized for individual images. Single-VP methods, such as \cite{bahng2022exploring}, introduce VP parameters as frames around input images, whereas \cite{chen2023understanding, tsao2024autovp} place these parameters around resized images. Diverse-VP methods include clustering-based approaches by \cite{huang2023diversity}, where cluster-specific prompts partially address the limitations of single prompts but fail to capture finer image-specific details. \cite{caisample} introduces auxiliary networks to generate individualized prompts; however, training separate auxiliary networks on small datasets increases the risk of overfitting \cite{yosinski2014transferable, zhang2021understanding}. Adapting outputs to downstream tasks necessitates a transformation because pre-trained model logits remain associated with the original source domain. Techniques such as random label mapping (RLM) align source labels to target labels. Frequency-based label mapping (FLM), proposed by \cite{tsai2020reprogramming}, leverages label distribution statistics, whereas semantic-based label mapping by \cite{yang-etal-2023-prompt} aligns classes based on semantic meanings. Iterative frequency-based label mapping (ILM), developed by \cite{chen2023understanding}, refines FLM iteratively. To further enhance adaptability, \cite{huang2023diversity} explores classifier head tuning, similar to linear probing (LP), to adapt the classifier head for specific tasks. Finally, \cite{tsao2024autovp} employs full mapping (FM) alongside an automated system to select optimal label mappings.

\subsection{Transfer Learning}
Transfer learning is a foundational technique in NLP and CV, allowing models to utilize knowledge obtained from pre-trained tasks to enhance performance on new, related tasks. In NLP, models such as BERT and GPT are first pre-trained on large-scale text datasets and subsequently fine-tuned for specific applications like sentiment analysis and question answering \cite{devlin2018bert, radford2019language}. Likewise, in CV, architectures including ResNet \cite{he2016deep}, and Vision Transformer \cite{dosovitskiy2020image, liu2021swin} are pre-trained on extensive datasets such as ImageNet \cite{deng2009imagenet}, and then adapted to tasks like object detection and image segmentation \cite{ren2016faster, he2017mask,li2025human}. Traditionally, transfer learning involves fully fine-tuning all parameters of the model on new task-specific data, using pre-trained weights as initialization. However, as models grow larger, this conventional approach becomes inefficient, leading to the development of parameter-efficient fine-tuning (PEFT) methods \cite{bahng2022exploring, tsao2024autovp, hu2021lora, pfeiffer2020adapterhub, li-liang-2021-prefix,jin2024learning}. These methods adjust only a subset of parameters or add task-specific modules while keeping most of the model unchanged, thereby reducing computational requirements and minimizing the risk of overfitting.

\vspace{-1mm}
\section{Methodology}\label{section_method}
\vspace{-1mm}
We propose adaptive Energy-Shaped Visual Prompting (\ours), a novel VP approach that integrates low-rank prompt initialization with an image-specific adaptation to enhance generalization and improve the discriminative performance of pre-trained models. An overview of \ours is illustrated in Figure \ref{figure_our_method}. We first formally define the visual prompting problem, then describe the low-rank initialization of VP, followed by the energy-shaped dynamic adaptation. Finally, we detail the optimization and inference procedures and present the comprehensive algorithm for \ours.

\subsection{Problem Statement}
\paragraph{Visual Prompt.} Consider a downstream target image dataset \( \mathcal{D} = \{(\mathbf{x}_1, y_1), \dots, \\(\mathbf{x}_n, y_n)\} \) with color channels \( c \) (typically 3) and a pre-trained vision model \( f_{\boldsymbol{\theta}} \) that processes images with resolution \( L \times L \) (set to \(224 \times 224\) in our experiments). Visual prompting begins by resizing input images to a specific size \( s \), as previous studies such as AutoVP \cite{tsao2024autovp} and ILM-VP \cite{chen2023understanding} have shown that the choice of \( s \) significantly affects VP performance. Task-specific tunable parameter frames, known as visual prompts, are then added to the resized images to create prompted images. This procedure can be formally expressed as:
\begin{equation}
    \mathcal{P}(\mathbf{x}) = \text{Resize}_{s}(\mathbf{x}) + \underbrace{g(\boldsymbol{\delta}) \odot \mathbf{m}}_{\text{Visual Prompts}}, \quad \mathbf{x} \in \mathcal{D},
    \label{formula_vp}
\end{equation}
where \( \text{Resize}_{s}(\cdot) \) resizes image \( \mathbf{x} \) to \( s \times s \) (typically smaller than \( L \) in existing VP frameworks) and pads the resized image with zeros to match the dimensions \( L \times L \) of \( f_{\boldsymbol{\theta}} \). Here, \( \boldsymbol{\delta} \in \mathbb{R}^{c \times L \times L} \) denotes the tunable parameters within VPs, \( g \) is a transformation function applied to these parameters (e.g., a Sigmoid function mapping \( \boldsymbol{\delta} \) values between 0 and 1, matching the input image range), \( \odot \) denotes element-wise multiplication, and \( \mathbf{m} \in \{0, 1\}^{c \times L \times L} \) is a binary mask indicating the prompted region. For universal single-VP methods, \( \mathbf{m} \) remains constant, whereas, in diverse-VP methods, \( \mathbf{m} \) can vary depending on the specific image \( \mathbf{x} \) \cite{caisample}. An illustration of the current VP frameworks is presented in Figure \ref{figure_existing_method}.

\paragraph{Output Transformation.} 
The output logits of a pre-trained model \( f_{\boldsymbol{\theta}} \) typically remain within the source domain (e.g., ImageNet \cite{deng2009imagenet}). To adapt these predictions to the target labels of downstream tasks (e.g., CIFAR100 \cite{krizhevsky2009learning}), an output transformation \( \mathcal{M} \) is necessary to map the source logits \( f_{\boldsymbol{\theta}}(\mathcal{P}(\mathbf{x})) \) to downstream predictions. Several output transformation methods have been proposed in recent studies. For instance, \cite{chen2023understanding} proposes iterative label mapping (ILM), which iteratively aligns the source logits with downstream predictions by utilizing frequency statistics from the training dataset at each epoch's start. \cite{huang2023diversity} investigates head tuning, directly adapting the classifier head of the pre-trained model to the downstream task, similar to linear probing (LP). \cite{tsao2024autovp} introduces full mapping (FM), adding an extra fully connected layer after the original classifier head to convert source logits into predictions suitable for downstream classes. In \ours, we examine three output transformation settings, LP, FM, and ILM, to assess their respective impacts in our experiments.

\subsection{Low-Rank Initialization of Visual Prompts}
Inspired by the inherent low-rank structure observed in deep neural networks, suggesting that low-rank matrices can retain the expressive capacity of full-rank weights \cite{jin2025lorvp, hu2021lora}, we introduce two low-rank matrices, \(\mathbf{B} \in \mathbb{R}^{c \times L \times r}\) and \(\mathbf{A} \in \mathbb{R}^{c \times r \times L}\), where \(r \ll L\), for initializing VPs. The product of these matrices, \(\boldsymbol{\delta} = \mathbf{B} \cdot \mathbf{A}\), acts as a universal VP, effectively capturing global dataset information. To streamline the VP process, enhance scalability, and reduce the complexity of hyperparameter tuning, we set a fixed resizing factor \(s = L\) for all images and employ a universal mask \(\mathbf{m} = \{1\}^{c \times L \times L}\), ensuring that the prompt influences the entire image. This yields the following VP formulation:
\begin{align}
    \mathcal{P}(\mathbf{x}) &= \text{Resize}_{L}(\mathbf{x}) + \boldsymbol{\delta} = \text{Resize}_{L}(\mathbf{x}) + \mathbf{B} \cdot \mathbf{A}, \quad \mathbf{x} \in \mathcal{D},
    \label{formula_our_vp}
\end{align}
where \(\text{Resize}_{L}(\cdot)\) resizes input images \(\mathbf{x}\) to dimensions \(L \times L\), and \(\mathbf{B}\) and \(\mathbf{A}\) represent the initialized parameters of the VPs. Following \cite{hu2021lora}, we initialize \(\mathbf{B}\) with zeros and \(\mathbf{A}\) with random Gaussian values, resulting in \(\mathbf{B} \cdot \mathbf{A}\) equaling zero at the start of training.

The low-rank visual prompt initialization naturally generalizes traditional full-rank prompts; setting \(r = L\) approximately restores the expressiveness of full-rank prompts. Notably, even at very low ranks, such as \(r = 4\), the low-rank VP approach achieves performance comparable to full-rank VPs \cite{jin2025lorvp}.

After the low-rank VP is applied, the output transformation \(\mathcal{M}\) is used to convert the pre-trained model's output logits on the prompted image \(f_{\boldsymbol{\theta}}(\mathcal{P}(\mathbf{x}))\) into target logits, represented as \(\mathcal{M}(f_{\boldsymbol{\theta}}(\mathcal{P}(\mathbf{x})))\).

\subsection{Energy-Shaped Dynamic Adaptation}
\cite{grathwohl2019your} suggests that classifiers, traditionally outputting logits, can also be interpreted as energy-based models. \cite{liu2020energy} further propose that energy functions more effectively identify out-of-distribution data compared to logits, assigning lower energy scores to observed data and higher scores to unobserved data. These findings indicate that energy scores provide valuable information about sampled images, aiding their recognition by pre-trained models.

Following \cite{grathwohl2019your} and \cite{liu2020energy}, we define the energy function at an input \(\mathbf{x}\) using the LogSumExp(\(\cdot\)) of the classifier's logits. After applying the low-rank VP initialization, the energy function is defined as:
{\small \begin{align}
    E_{\boldsymbol{\theta},\boldsymbol{\delta}}(\mathbf{x})=-\text{LogSumExp}_y(\mathcal{M}(f_{\boldsymbol{\theta}}(\mathcal{P}(\mathbf{x})))[y])=-\log \sum_{y} \exp(\mathcal{M}(f_{\boldsymbol{\theta}}(\mathcal{P}(\mathbf{x})))[y]),
    \label{formula_energy_function}
\end{align}}
where \(E_{\boldsymbol{\theta},\boldsymbol{\delta}}(\mathbf{x})\) denotes the energy score and \(\mathcal{M}(f_{\boldsymbol{\theta}}(\mathcal{P}(\mathbf{x})))[y]\) refers to the logit of class \(y\).

Energy-based models \cite{lecun2006tutorial} posit that the probability density function \(d_{\boldsymbol{\theta}}(\mathcal{P}(\mathbf{x}))\) of a prompted image can be expressed as:
\begin{equation}
    d_{\boldsymbol{\theta}}(\mathcal{P}(\mathbf{x})) = \frac{\exp(-E_{\boldsymbol{\theta}}(\mathcal{P}(\mathbf{x})))}{Z(\boldsymbol{\theta})}
    \label{formula_energy_distribution}
\end{equation}
where \(d_{\boldsymbol{\theta}}(\cdot)\) is the probability density, \(E_{\boldsymbol{\theta}}(\cdot)\) is the energy function mapping images to scalar values, and \(Z(\boldsymbol{\theta}) = \int_{\mathcal{P}(\mathbf{x})} \exp(-E_{\boldsymbol{\theta}}(\mathcal{P}(\mathbf{x})))\) is the normalizing constant (partition function).

Taking the logarithm of both sides of Equation (\ref{formula_energy_distribution}), we obtain:
\begin{equation}
    \log d_{\boldsymbol{\theta}}(\mathcal{P}(\mathbf{x})) = -E_{\boldsymbol{\theta}}(\mathcal{P}(\mathbf{x})) - \log Z(\boldsymbol{\theta}),
    \label{formula_log_p}
\end{equation}
where \(\log\) denotes the natural logarithm. Equation (\ref{formula_log_p}) indicates that logits for a prompted image can be expressed through the energy function and a normalization constant.

By computing the gradients of logits with respect to the prompted image, we define our image-specific dynamic VP adaptation \(\boldsymbol{\mu}_{\boldsymbol{\delta}}(\mathbf{x})\) as follows:
\begin{align}
    \boldsymbol{\mu}_{\boldsymbol{\delta}}(\mathbf{x}) = \alpha \frac{\partial \log d_{\boldsymbol{\theta}}(\mathcal{P}(\mathbf{x}))}{\partial \mathcal{P}(\mathbf{x})} = -\alpha \frac{\partial E_{\boldsymbol{\theta},\boldsymbol{\delta}}(\mathbf{x})}{\partial \mathcal{P}(\mathbf{x})} = -\alpha \frac{\partial E_{\boldsymbol{\theta},\boldsymbol{\delta}}(\mathbf{x})}{\partial \boldsymbol{\delta}},
    \label{formula_image_specific_vp}
\end{align}where \(\alpha\) is a learnable \textbf{strength coefficient} controlling the adaptation strength.

By integrating the image-specific VP adaptation \(\boldsymbol{\mu}_{\boldsymbol{\delta}}(\mathbf{x})\) with the low-rank initialized VP \(\mathcal{P}(\mathbf{x})\), \ours dynamically adjusts VPs according to the specific features of each image. The dynamically adapted VP image is formulated as:
\begin{align}
    \mathcal{P}(\mathbf{x})+\boldsymbol{\mu}_{\boldsymbol{\delta}}(\mathbf{x}) = \mathcal{P}(\mathbf{x})-\alpha \frac{\partial E_{\boldsymbol{\theta},\boldsymbol{\delta}}(\mathbf{x})}{\partial \boldsymbol{\delta}}.
    \label{formula_final_vp}
\end{align}
Equation (\ref{formula_final_vp}) allows image-specific VPs without requiring labels, additional parameters, or auxiliary networks, thus enhancing parameter efficiency and reducing overfitting risk. Additionally, it integrates flexibly with various pre-trained models, improving scalability compared to methods such as DAM-VP \cite{huang2023diversity} and SMM \cite{caisample}, which rely on clustering or auxiliary networks. Our experiments (Section \ref{section_ood_performance}) further demonstrate the improved OOD performance using the energy function \cite{grathwohl2019your, liu2020energy}.

\subsection{Meta-Learning of Low-Rank Prior and Dynamic Adaptation}
We propose a meta-learning framework that jointly optimizes (i) image-specific dynamic prompt adaptation in the inner loop and (ii) a global low-rank prompt prior in the outer loop. Through meta-learning, this paradigm establishes a mutually reinforcing relationship: a well-optimized low-rank prior facilitates precise dynamic adaptation, which conversely provides optimal feedback to continuously improve the prior. This synergy ensures fast convergence and yields highly effective per-image prompts \emph{without auxiliary networks or additional trainable modules}. By avoiding the heavy parameterizations and complex designs of existing diverse-VP methods (e.g., DAM-VP), it also addresses their scalability bottlenecks while ensuring simplicity for real-world deployment.

Specifically, the optimization procedure of \ours adopts this meta-learning approach \cite{finn2017model, andrychowicz2016learning, guan2025meta} through two nested stages:

\textbf{Inner Loop (Dynamic Adaptation):} For each image $\mathbf{x}$ in a sampled batch, given the current global low-rank VP prior $\boldsymbol{\delta}$, we compute the dynamic adaptation term $\boldsymbol{\mu}_{\boldsymbol{\delta}}(\mathbf{x})$ using the energy-based gradient as defined in Equation (\ref{formula_image_specific_vp}). This generates the image-specific prompted input $\mathcal{P}(\mathbf{x}) + \boldsymbol{\mu}_{\boldsymbol{\delta}}(\mathbf{x})$ instantly, without requiring label supervision or optimizing auxiliary network parameters.

\textbf{Outer Loop (Prior Optimization):} We evaluate the downstream target loss using the dynamically adapted image-specific prompts. Subsequently, we update the global low-rank prompt prior parameters $\boldsymbol{\delta}$, the adaptation strength coefficient $\alpha$, and the output transformation $\mathcal{M}$ via gradient descent:
\begin{equation}
    \underset{\boldsymbol{\delta}, \alpha, \mathcal{M}}{\text{minimize}}\quad \mathbb{E}_{(\mathbf{x}, y)\in \mathcal{D}} \mathcal{L}(\mathcal{M}(f_{\boldsymbol{\theta}}(\mathcal{P}(\mathbf{x}) + \boldsymbol{\mu}_{\boldsymbol{\delta}}(\mathbf{x}))), y).
    \label{formula_optimization}
\end{equation}

The pseudo-code detailing this procedure is provided in Algorithm \ref{algorithm_dvp} for clarity and reference. Upon completion of training, the optimal global low-rank prior and strength coefficient $\alpha$ are determined. During inference, these meta-learned configurations are directly applied. For any unseen image, its dynamic VP adaptation is computed on the fly according to Equation (\ref{formula_final_vp}), yielding a tailored visual prompt for downstream tasks.

\begin{algorithm}[t]
   \caption{Energy-Shaped Visual Prompting (\ours)}
   \label{algorithm_dvp}
\begin{algorithmic}
   \State {\bfseries Input:} Pre-trained model $f_{\btheta}$, Image Dataset $\mathcal{D}$, Initialize output transformation $\mathcal{M}$, Initialize learnable image-specific prompt coefficient $\alpha$, Initialize the global low-rank VP prior $\bdelta = \bB \cdot \bA$, where $\bB \in \mathbb{R}^{c \times L \times r}$ and $\bA \in \mathbb{R}^{c \times r \times L}$, $r \ll L$
   \Repeat \ \textbf{(Outer Loop)}
       \State Sample a batch of data $\mathcal{B} \subset \mathcal{D}$
       \For{each $(\bx, y) \in \mathcal{B}$} \ \textbf{(Inner Loop)}
           \State Compute low-rank initialization prompted image $\mathcal{P}(\bx) = \text{Resize}_{L}(\bx) + \bdelta$ 
           \State Compute energy-based dynamic adaptation $\bmu_{\bdelta}(\bx) = -\alpha \frac{\partial E_{\btheta,\bdelta}(\bx)}{\partial \bdelta}$
           \State Formulate dynamically adapted prompted image $\mathcal{P}(\bx) + \bmu_{\bdelta}(\bx)$
       \EndFor
       \State Compute the meta-objective loss $ \mathcal{L}_{target} = \mathbb{E}_{(\bx, y)\in \mathcal{B}}\mathcal{L}(\mathcal{M}(f_{\btheta}(\bx_{prompted})), y)$
       \State Update global prior $\bdelta$, strength $\alpha$, and $\mathcal{M}$ using gradient $\nabla_{\bdelta, \alpha, \mathcal{M}} \mathcal{L}_{target}$
   \Until Convergence
\end{algorithmic}
\end{algorithm}

\vspace{-1mm}
\section{Experiment}\label{section_experiment}
\vspace{-1mm}
To evaluate the effectiveness and efficiency of \ours, we adopt the standard VP evaluation protocol, assessing models pre-trained on large-scale datasets across various visual domains. Consistent with prior studies such as AutoVP \cite{tsao2024autovp} and SMM \cite{caisample}, we examine the in-distribution performance of \ours on ten small-scale image classification datasets. To investigate out-of-distribution (OOD) performance, we further analyze models pre-trained on ImageNet-21K \cite{deng2009imagenet} and adapted with VPs on ImageNet-1K. Furthermore, we perform extensive empirical evaluations focusing on the following aspects: (1) demonstrating the superior in-distribution performance of \ours across multiple datasets and architectures; (2) assessing the OOD robustness of \ours on ImageNet variants; and (3) performing comprehensive ablation studies and detailed analyses to investigate the effects of the energy-based adaptation method, the strength coefficient $\alpha$, and the overall efficiency of \ours.

\subsection{Implementation Details}\label{section_implementation_details}
\paragraph{Networks.} We employ five architectures for our experiments, all operating at a resolution of $224 \times 224$: (1) ResNet-18 and ResNet-50 \cite{he2016deep}, pre-trained on ImageNet-1K, each with a classifier head for 1000 classes; (2) ViT-B/32 \cite{dosovitskiy2020image}, pre-trained on ImageNet-21K and fine-tuned on ImageNet-1K, also with a 1000-class classifier head; (3) Swin-B \cite{liu2021swin}, pre-trained on ImageNet-21K, with a classifier head for 21,841 classes; (4) CLIP \cite{radford2021learning}, a vision-language model employing a ViT-B/32 architecture as its vision encoder. The weights are publicly available through the official PyTorch Model Zoo\footnote{\href{https://pytorch.org/vision/stable/models.html}{https://pytorch.org/vision/stable/models.html}} and the Hugging Face Timm Library\footnote{\href{https://huggingface.co/models?library=timm}{https://huggingface.co/models?library=timm}}.

\paragraph{Datasets.} For pre-training, we use the ImageNet-1K dataset \cite{deng2009imagenet}, containing 1K classes and 1.3M images, and the ImageNet-21K dataset \cite{deng2009imagenet}, which includes 21K classes and 14M images. We evaluate the in-distribution performance of \ours on ten small datasets: Tiny ImageNet \cite{le2015tiny}, EuroSAT \cite{helber2019eurosat}, OxfordPets \cite{parkhi2012cats}, Food101 \cite{bossard2014food}, DTD \cite{cimpoi2014describing}, Flowers102 \cite{nilsback2008automated}, CIFAR10/100 \cite{krizhevsky2009learning}, SVHN \cite{netzer2011reading}, and GTSRB \cite{houben2013detection}. To assess the OOD robustness of \ours, we perform experiments by training on ImageNet-1K and evaluating on ImageNet-R \cite{hendrycks2021many}, ImageNet-Sketch \cite{wang2019learning}, ImageNet-A \cite{hendrycks2021natural}, and ImageNet-V2 \cite{recht2019imagenet}.

\paragraph{Baselines.} We select six representative SOTA methods as baselines for comparison: (1) \textit{ILM-VP} \cite{chen2023understanding}, a Single-VP method that analyzes the effect of frequency-based label mapping (FLM) and introduces iterative label mapping (ILM) for improved performance; (2) \textit{AutoVP} \cite{tsao2024autovp}, a leading Single-VP method that automatically selects VP configurations, including prompt sizes and label mapping (LM) strategies; (3) \textit{DAM-VP} \cite{huang2023diversity}, a Diverse-VP method that clusters downstream datasets into smaller homogeneous subsets, each optimized separately with its own prompt; (4) \textit{SMM} \cite{caisample}, another Diverse-VP method that uses a lightweight convolutional neural network (CNN) and patch-wise interpolation to generate image-specific three-channel masks instead of a shared, predefined mask; (5) \textit{LoR-VP} \cite{jin2025lorvp}, a SOTA Single-VP method that utilizes a global low-rank VP and (6) \textit{LP} \cite{iofinova2022well}, which modifies the classifier head of the pre-trained model to adapt to downstream tasks, a common technique in transfer learning.

\paragraph{Training and Evaluation.} Results for baseline methods, including ILM-VP, AutoVP, DAM-VP, SMM, and LoR-VP, are reproduced following the experimental configurations specified in their original papers. For \ours, we use LP as the default output transformation, resize all input images to \(224 \times 224\), and set the rank in the low-rank initialization design to \(4\), resulting in only approximately 5K trainable parameters. Optimal hyperparameters for \ours are determined through a grid search. We set the learning rate to \(0.001\), the number of training epochs to \(20\), and initialize the strength coefficient at \(0.05\). The experiments are conducted on NVIDIA Quadro RTX6000 GPUs with 24GB of memory, and the results reported are averaged over three independent runs.

\vspace{-1mm}
\section{Main Results}
\vspace{-1mm}
\paragraph{In-distribution Performance.} To evaluate the in-distribution generalization performance of \ours, we assess its effectiveness across ten downstream datasets, including natural and artificial objects, scenes, textures, and various original image sizes, following \cite{tsao2024autovp} and \cite{caisample}. The quantitative results of using ImageNet-1K pre-trained ResNet-18 and ImageNet-21K pre-trained and ImageNet-1K fine-tuned ViT-B/32 are shown in Table \ref{table_cnn_in_distribution}, we observe that: \ours outperforms all baselines across all networks on ten datasets, achieving an average accuracy improvement of \textbf{4.5\%} over the Single-VP method AutoVP and \textbf{1.6\%} over the Diverse-VP method DAM-VP using ResNet-18. Notably, \ours accomplishes this while utilizing $\times$ 12 fewer VP parameters than AutoVP and $\times$ 340 fewer than DAM-VP. These results demonstrate the effectiveness and parameter efficiency of the dynamic VPs attained by the introduced energy method in \ours.

\begin{table*}[t]
\centering
\caption{\textbf{Performance of ImageNet-1K pre-trained ResNet-18 and  ImageNet-21K pre-trained ViT-B/32  on downstream datasets.} Overview of the performance of \ours compared to five SOTA baseline methods. Results are averaged over three runs. \ours outperforms all baselines across models and datasets.}
\label{table_cnn_in_distribution}
\renewcommand{\arraystretch}{1.0}
\resizebox{1\textwidth}{!}{
\begin{tabular}{c|c|cccccccccc|c}
\toprule
\textbf{Model} & \textbf{Method} & \textbf{Tiny-ImageNet} & \textbf{EuroSAT} & \textbf{OxfordPets} & \textbf{ Food101} & \textbf{DTD} & \textbf{Flowers102} & \textbf{ CIFAR10} & \textbf{ CIFAR100} & \textbf{SVHN} & \textbf{GTSRB} & \textbf{Average} \\
\toprule
\multirow{7}{*}{ResNet-18} & LP \textcolor{gray}{[CVPR22]} & 65.12 & 93.56 & 87.25 & 51.24 & 59.88 & 88.12 & 87.05 & 67.01 & 65.56 & 78.02 & 74.28 \\ 
 & ILM-VP \textcolor{gray}{[CVPR23]} & 14.11 & 85.12 & 65.32 & 24.94 & 35.23 & 27.85 & 66.31 & 25.32 & 75.12 & 53.05 & 47.24 \\  
 & AutoVP \textcolor{gray}{[ICLR24]} & 59.72 & 93.11 & 82.77 & 54.24 & 54.67 & 85.32 & 87.55 & 63.75 & 83.89 & 81.73 & 74.68 \\ 
 & DAM-VP \textcolor{gray}{[CVPR23]} & 67.15 & \textbf{94.45} & 87.13 & 54.01 & 61.29 & \textbf{89.51} & 87.88 & 68.82	& 84.35 & 81.52	& 77.61 \\  
 & SMM \textcolor{gray}{[ICML24]} & 39.81 & 92.15 & 74.01 & 17.78 & 33.56 & 38.84 & 72.56 & 38.89 & 84.14 & 80.56 & 57.23 \\  
  & LoR-VP \textcolor{gray}{[ICLR25]} & 68.28 & 92.54 & 83.31 & 54.78 & 61.21 & 87.22 & 88.64 & 69.88 & 83.14 & 81.36 & 77.04 \\ 
\cmidrule{2-13}
 & \ours (Ours) & \textbf{68.89} & 94.42 &\textbf{88.63} & \textbf{55.67} & \textbf{66.62} & 89.35 & \textbf{89.63} & \textbf{71.15} & \textbf{84.84} & \textbf{82.58} & \textbf{79.18} \\ 
\midrule
\multirow{7}{*}{ViT-B/32} & LP \textcolor{gray}{[CVPR22]} & 83.88 & 96.06 & 92.03 & 82.64 & 70.12 & 98.23 & 96.54 & 86.01 & 85.31 & 87.77 & 87.86 \\ 
 & ILM-VP \textcolor{gray}{[CVPR23]} & 32.38 & 88.24 & 78.90 & 48.28 & 42.64 & 64.22 & 85.23 & 40.02 & 80.85 & 67.89 & 62.87 \\
 & AutoVP \textcolor{gray}{[ICLR24]} & 82.39 & 95.65 & 92.12 & 81.81 & 70.65 & 98.35 & 95.42 & 85.86 & 85.01 & 87.12 & 87.44 \\
 & DAM-VP \textcolor{gray}{[CVPR23]} & 84.90 & 95.72 & 92.40 & 86.77 & 73.08 & \textbf{99.21} & 97.36 & 88.10 & 87.84 & 90.62 & 89.60 \\
 & SMM \textcolor{gray}{[ICML24]} & 79.72 & 93.37 & 83.91 & 64.68 & 45.42 & 79.31 & 97.35 & 82.67 & 89.55 & 80.71 & 79.67 \\ 
 & LoR-VP \textcolor{gray}{[ICLR25]} & 85.85 & 96.25 & 92.18 & 83.51 &  72.30 & 98.58 & 97.52 & 88.65 & 86.31 & 88.07 & 88.92 \\ 
\cmidrule{2-13}
 & \ours (Ours)  & \textbf{86.65} & \textbf{96.45} & \textbf{92.48} & \textbf{87.82} & \textbf{74.51} & 99.10 & \textbf{97.81} & \textbf{89.44} & \textbf{90.92} & \textbf{91.78} & \textbf{90.70} \\
\bottomrule
\end{tabular}}
\end{table*}

\begin{figure*}[!ht]
  \centering
  \begin{subfigure}{0.24\textwidth}
    \includegraphics[width=\linewidth]{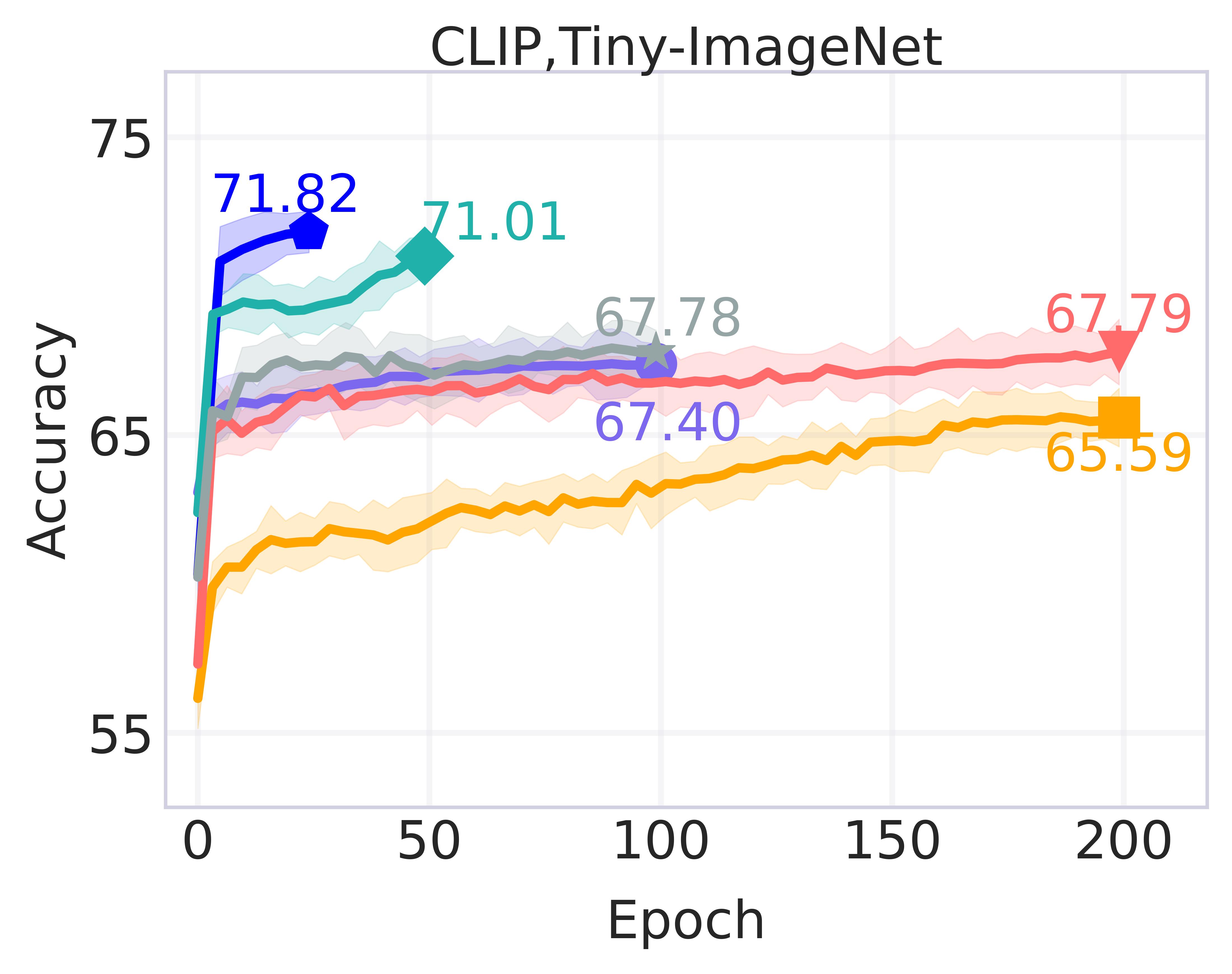}
  \end{subfigure}
  \hfill
  \begin{subfigure}{0.24\textwidth}
    \includegraphics[width=\linewidth]{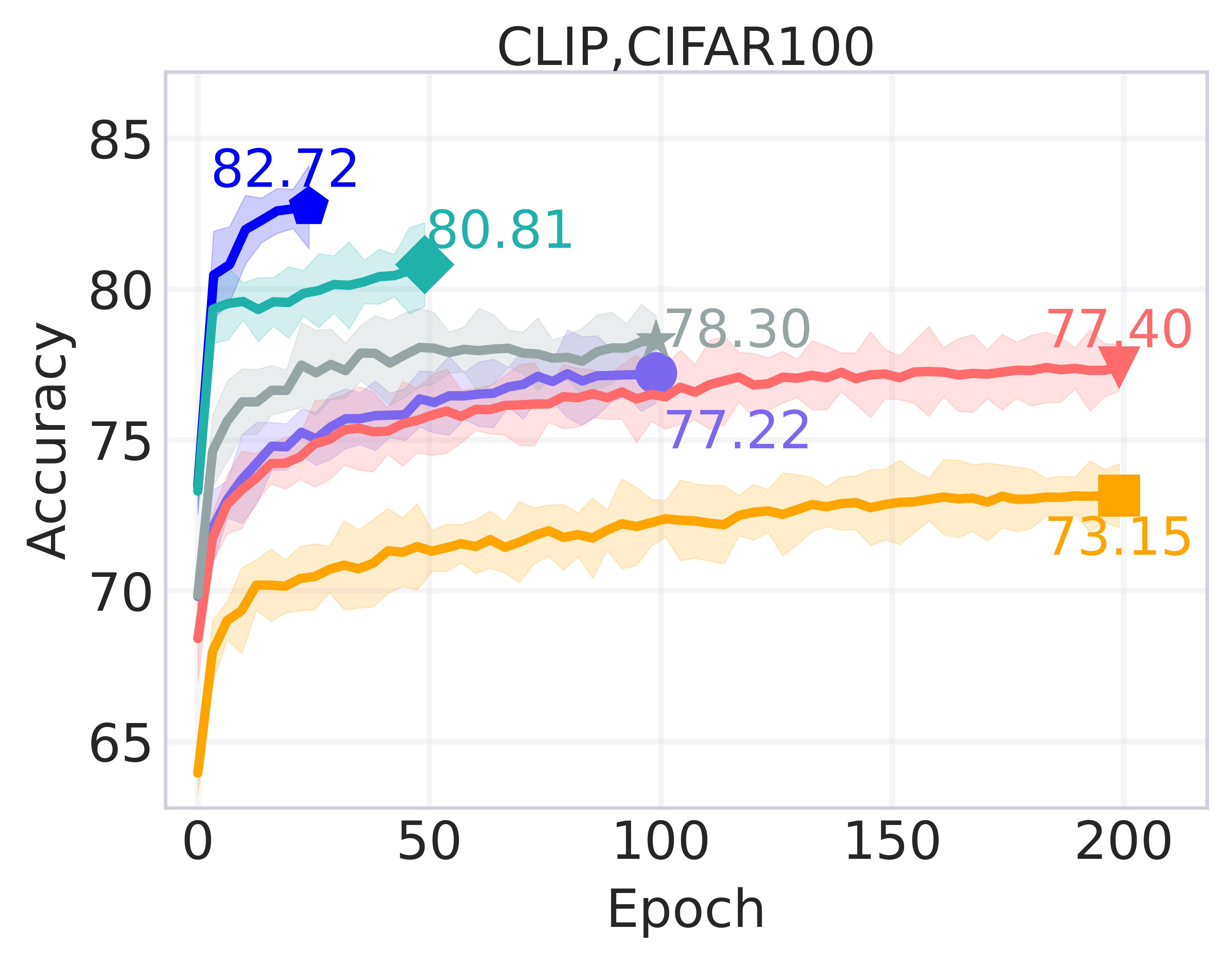}
  \end{subfigure}
  \hfill
  \begin{subfigure}{0.24\textwidth}
    \includegraphics[width=\linewidth]{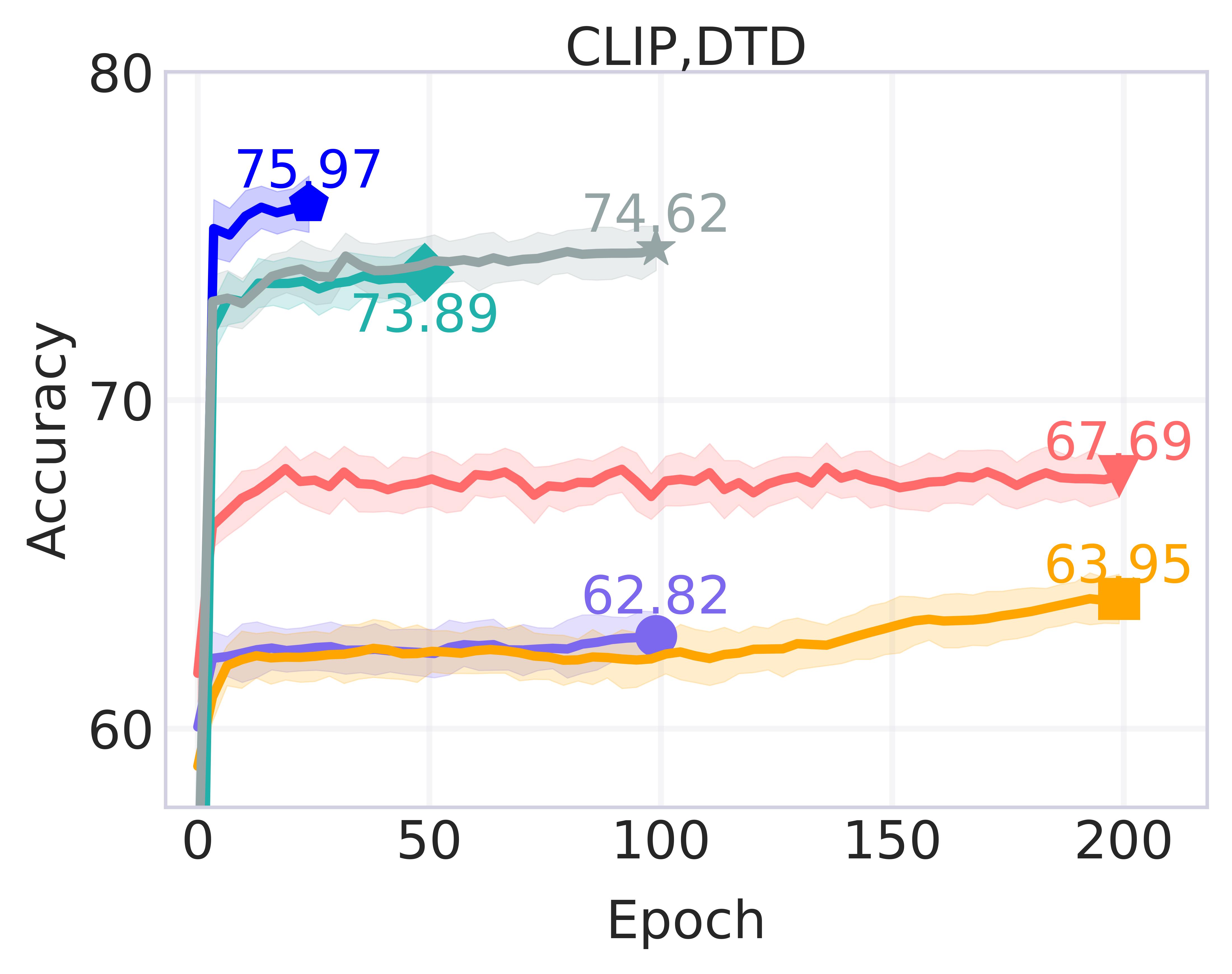}
  \end{subfigure}
  \hfill
  \begin{subfigure}{0.24\textwidth}
    \includegraphics[width=\linewidth]{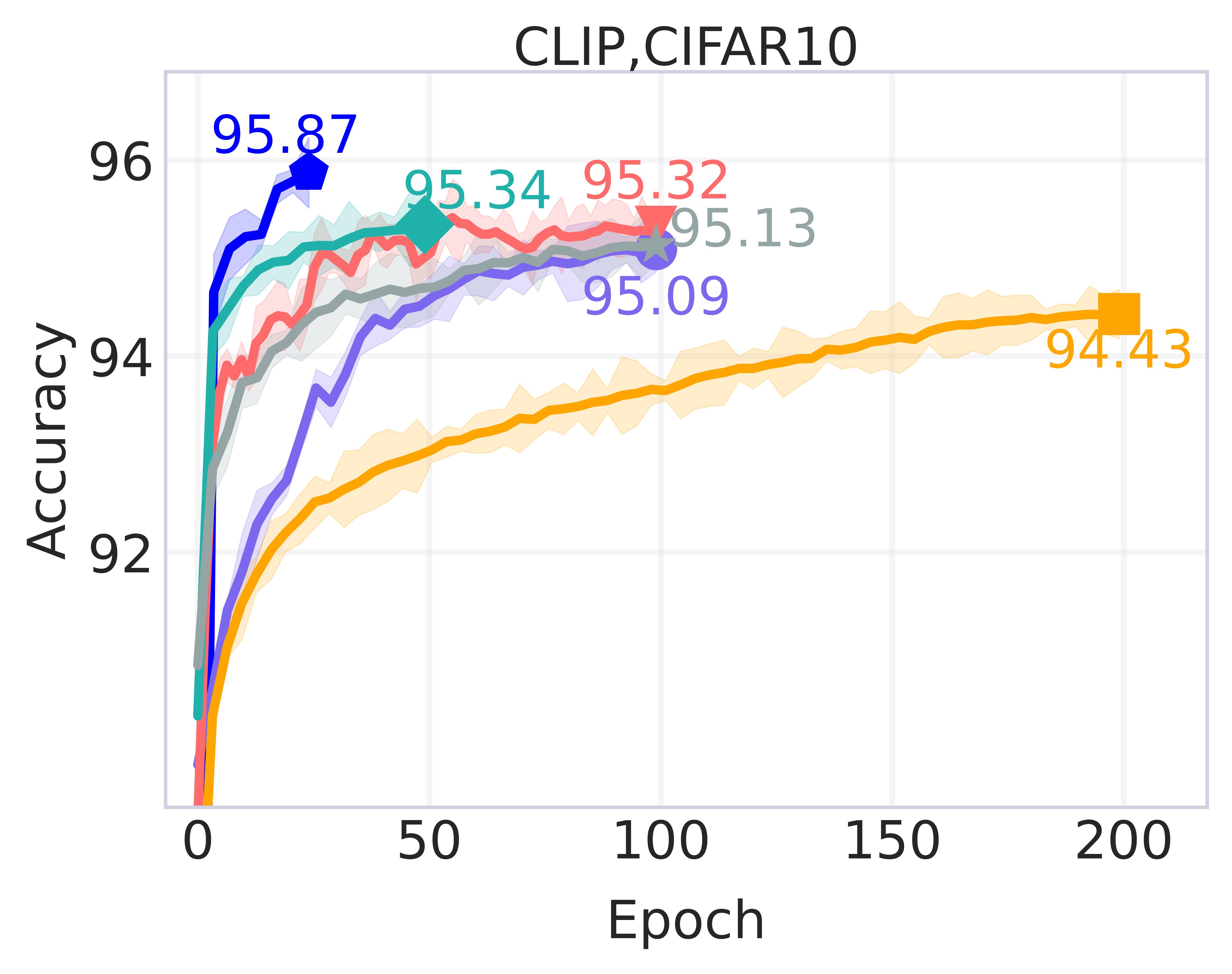}
  \end{subfigure}
  
  \begin{subfigure}{0.24\textwidth}
    \includegraphics[width=\linewidth]{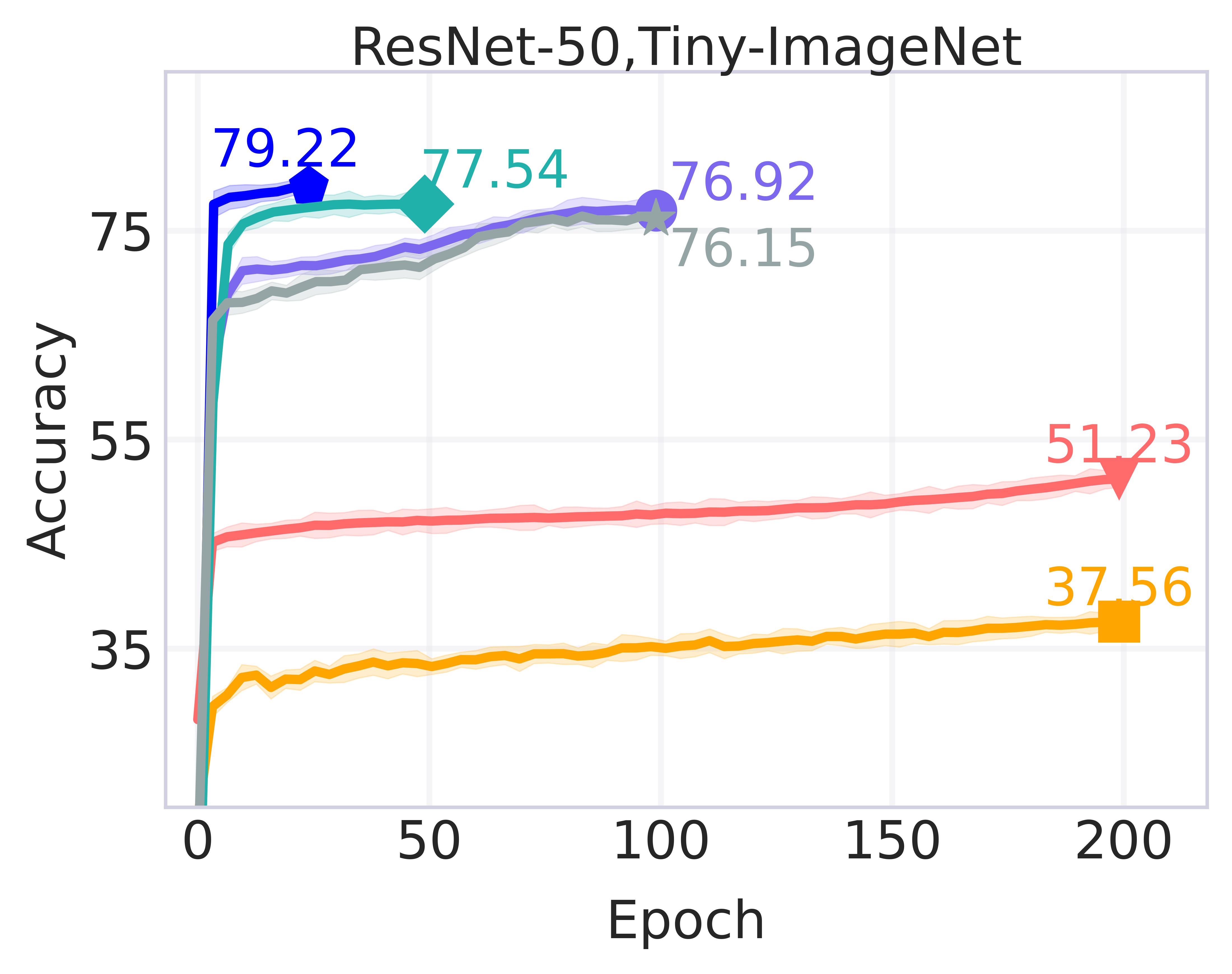}
  \end{subfigure}
  \hfill
  \begin{subfigure}{0.24\textwidth}
    \includegraphics[width=\linewidth]{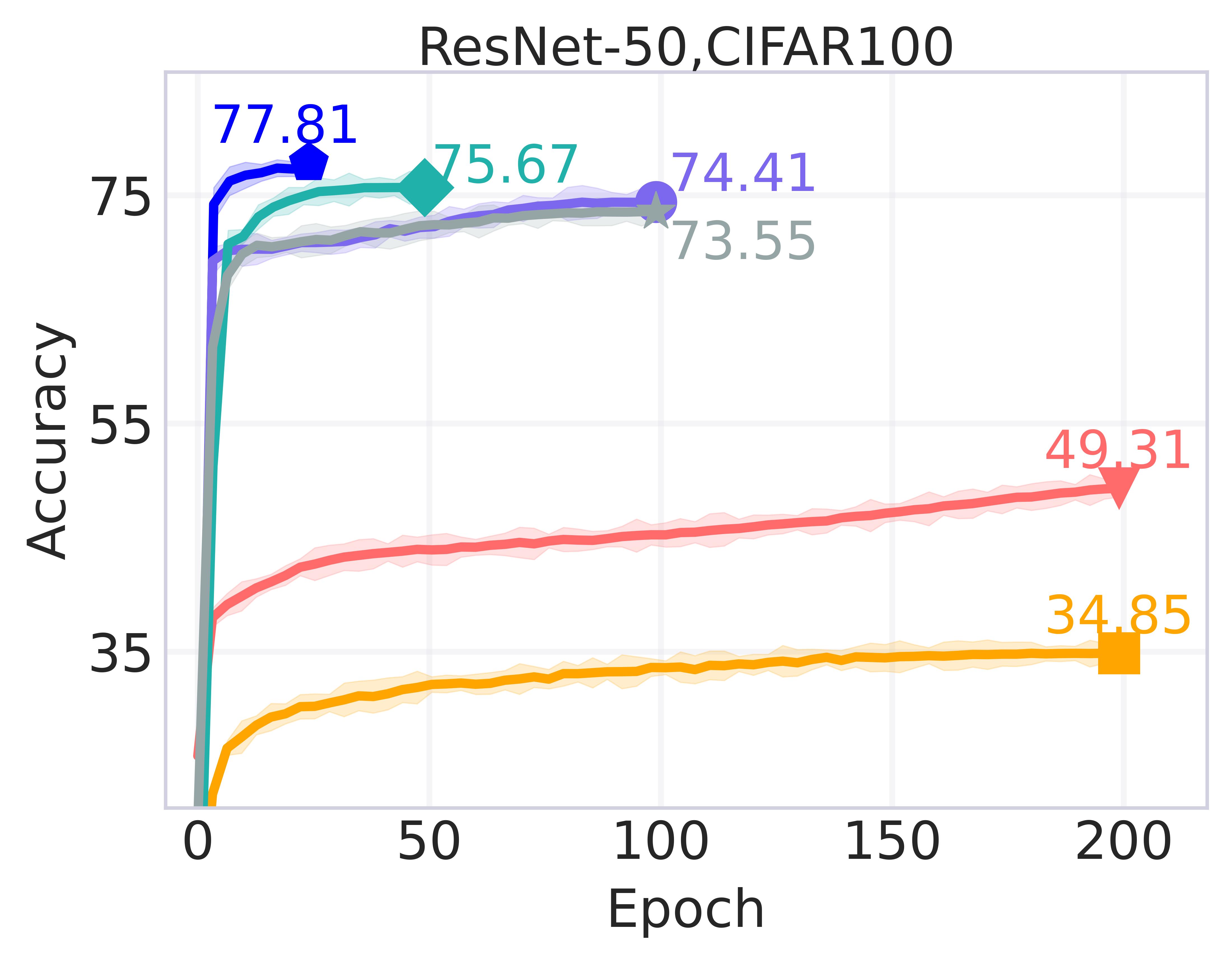}
  \end{subfigure}
  \hfill
  \begin{subfigure}{0.24\textwidth}
    \includegraphics[width=\linewidth]{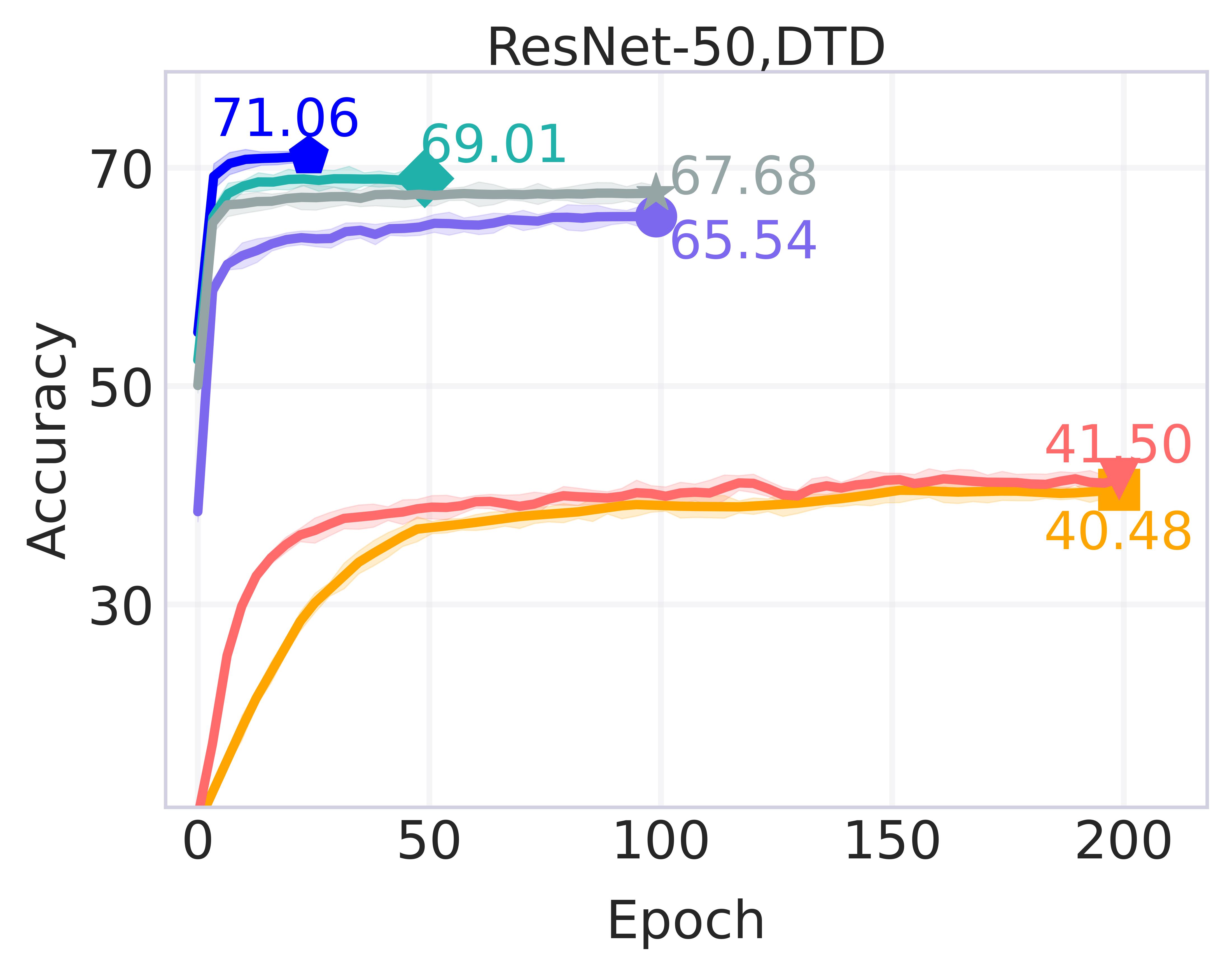}
  \end{subfigure}
  \hfill
  \begin{subfigure}{0.24\textwidth}
    \includegraphics[width=\linewidth]{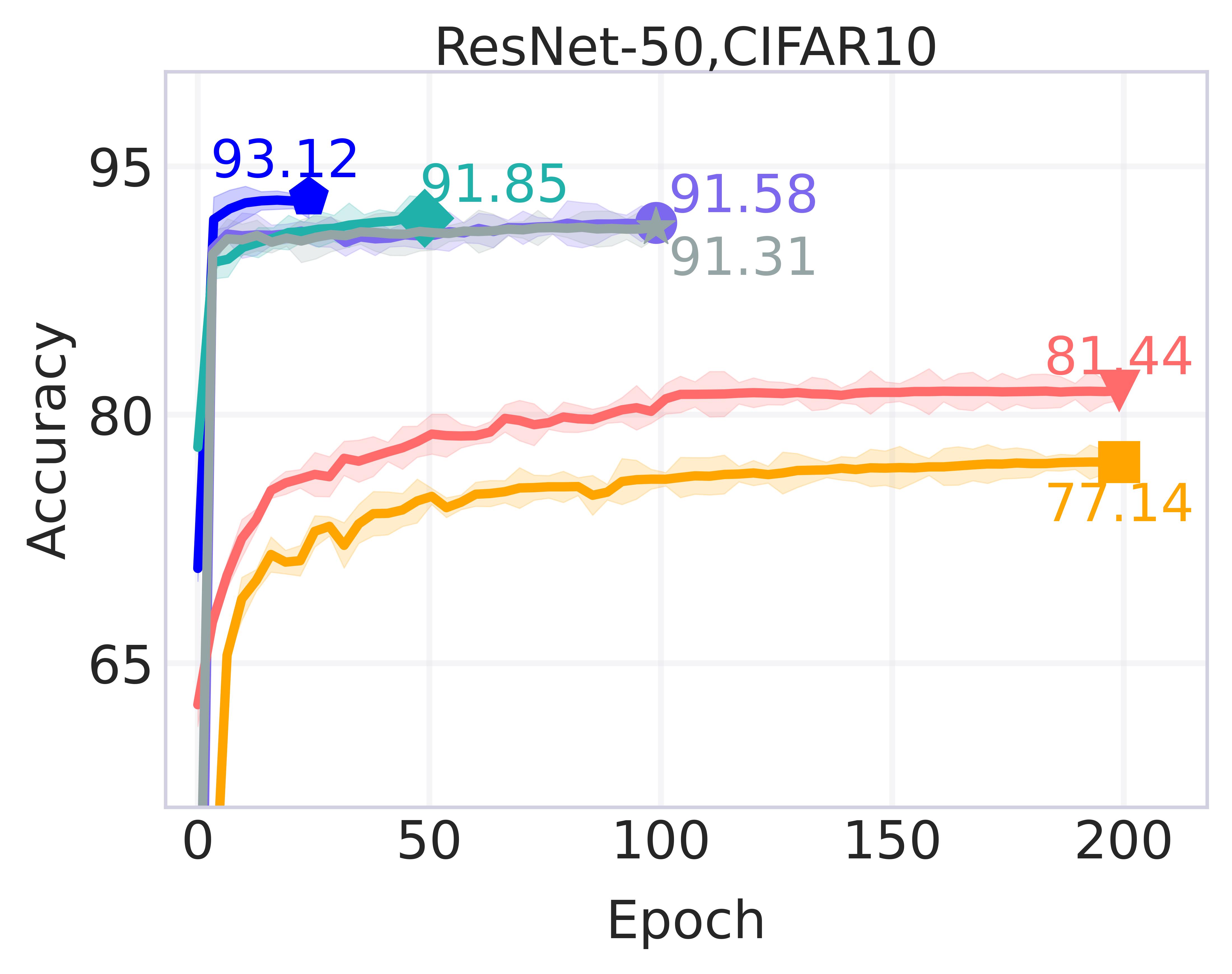}
  \end{subfigure}
  \begin{subfigure}{0.65\textwidth}
    \includegraphics[width=\linewidth]{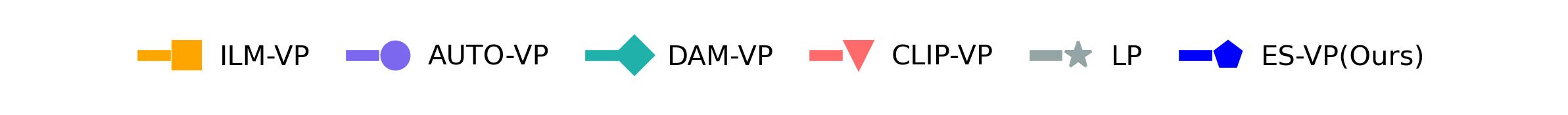}
  \end{subfigure}
  \caption{\textbf{Performance of ImageNet-1K pre-trained ResNet-50 and CLIP Pre-trained Models on Downstream Datasets.} Overview of the performance of \ours compared to five baseline methods. The final performance of each method is indicated by the markers. \ours consistently outperforms all baselines across various models and datasets with fewer training epochs.}
  \label{Figure_performance_resnet50_vit}
\end{figure*}

To visualize the training process of visual prompting and provide qualitative results, we show the test accuracy of each epoch during training using CLIP with ViT-B/32 as the vision encoder and ImageNet-1K pre-trained ResNet-50 on four datasets in Fig. \ref{Figure_performance_resnet50_vit}. We can draw the following positive observations: \ding{182} \ours consistently achieves the best final accuracy across all model and dataset combinations; significantly, \ours achieves $\textbf{8.0\%}$ higher accuracy than the Diverse-VP method DAM-VP using CLIP on DTD and $\textbf{28.2\%}$ higher accuracy than the Diverse-VP method SMM on ResNet-50 and CIFAR100. \ding{183} \ours converges the fastest among all VP methods, achieving better performance under fewer training epochs, highlighting the training efficiency of \ours.

\begin{table}[t]
    \centering
    \begin{minipage}[t]{0.54\textwidth}
        \centering
        \caption{\textbf{Out-of-Distribution Generalization Performance.} Evaluation of OOD generalization performance using ImageNet-21K pre-trained Swin-B trained on ImageNet-1K, tested across four OOD datasets.}
        \label{table_out_of_distribution_performance}
        \resizebox{\linewidth}{!}{
        \begin{tabular}{r|c|cccc}
        \toprule
        {\centering \multirow{2}{*}{Method    }} & Source & \multicolumn{4}{c}{Target} \\ 
        \cmidrule(lr){2-2}\cmidrule(lr){3-6}
         & ImageNet-1K & -R & -Sketch & -A & -V2 \\
        \midrule
        LP \textcolor{gray}{[CVPR22]} & 83.47 & 51.24 & 39.96 & 27.38 & 71.73 \\
        AutoVP \textcolor{gray}{[ICLR24]} & 78.98 & 38.14 & 28.89 & 17.91 & 67.38 \\
        DAM-VP \textcolor{gray}{[CVPR23]} & 83.97 & 52.08 & 41.12 & 27.90 & 72.43 \\
        LoR-VP \textcolor{gray}{[ICLR25]}  & 84.04 & 52.27 & 41.13 & 27.89 & 72.38 \\
        \midrule
        \ours (Ours) & \textbf{84.17} & \textbf{53.22} & \textbf{42.08} & \textbf{28.41} & \textbf{74.05} \\
        \bottomrule
        \end{tabular}}
    \end{minipage}%
    \hfill
    \begin{minipage}[t]{0.44\textwidth}
        \centering
        \caption{\textbf{Ablation Study on Components of \ours.} Results obtained using ImageNet-21K pre-trained ViT-B/32 and ImageNet-1K pre-trained ResNet-18 on the DTD dataset.}
        \label{table_ablation_components}
        \resizebox{\linewidth}{!}{
        \begin{tabular}{lcc}
        \toprule
        Configuration & ViT-B/32 & ResNet-18 \\
        \midrule
        Baseline & 70.12 & 59.88 \\
        + Low-Rank Init. & 72.30 & 61.21 \\
        + Dynamic Adaptation & 72.92 & 64.68 \\
        + Both Components & \textbf{74.51} & \textbf{66.62} \\
        \bottomrule
        \end{tabular}
        }
    \end{minipage}
\end{table}

\paragraph{Out-of-distribution Performance.}\label{section_ood_performance} A key advantage of \ours is the use of an energy-based method to generate dynamic VPs, which helps in identifying OOD data \cite{grathwohl2019your, liu2020energy}. To evaluate the OOD robustness and potential practical applicability of \ours, we perform experiments using ImageNet-21K pre-trained Swin-B trained further on ImageNet-1K. We subsequently assess the resulting model and VPs on four OOD datasets. The performance comparison between \ours and the 3 strongest baseline methods is presented in Table Table \ref{table_out_of_distribution_performance}. Notably, \ding{182} \ours achieves the highest performance among all baselines on the extensive ImageNet-1K dataset, further validating its effectiveness on large-scale datasets. \ding{183} \ours consistently demonstrates superior OOD robustness and generalization capabilities across all four OOD datasets, showing an average accuracy improvement of $\textbf{11.4\%}$ over AutoVP and \(1.1\%\) over DAM-VP. These results highlight the robustness of \ours and its potential advantages in complex, real-world applications.

\vspace{-1mm}
\section{Additional Investigation}
\vspace{-1mm}
\paragraph{Impact of Components in \ours.}
We conduct ablation studies on the two primary components of \ours: the low-rank prompt initialization and the energy-based dynamic adaptation. The experiments are performed using ImageNet-21K pre-trained ViT-B/32 and ImageNet-1K pre-trained ResNet-50, evaluated on the DTD dataset. Four distinct configurations are studied: (1) Without low-rank initialization and dynamic adaptation, investigating the effect of solely applying LP as the output transformation. (2) With low-rank initialization but without dynamic adaptation, evaluating the impact of only using low-rank initialized VP. (3) With dynamic adaptation but without low-rank initialization, examining the sole contribution of the energy-based adaptation. In this setting, \(\boldsymbol{\delta} = 0\), according to the final VP formulation in  (\ref{formula_final_vp_0}), the dynamically prompted image can be formulated as:
\begin{align}
    \mathcal{P}(\mathbf{x}) + \boldsymbol{\mu}_{\boldsymbol{\delta}}(\mathbf{x}) = \text{Resize}_{L}(\mathbf{x}) - \alpha \frac{\partial E_{\boldsymbol{\theta}}(\mathbf{x})}{\partial \boldsymbol{\bx}}.
    \label{formula_final_vp_0}
\end{align}
(4) With both low-rank initialization and dynamic adaptation, representing the complete \ours method. Experimental results presented in Table \ref{table_ablation_components} allow the following observations: \ding{182} Dynamic adaptation and low-rank prompt initialization are orthogonal, as the highest performance is achieved when both components are combined. \ding{183} Dynamic adaptation consistently enhances performance, whether or not low-rank initialization is used. \ding{184} Dynamic adaptation provides greater improvement than low-rank initialization alone; for instance, using ResNet-18, dynamic adaptation yields a \(4.8\%\) accuracy improvement over setting (1), whereas low-rank initialization contributes a \(1.3\%\) improvement.

\paragraph{Impact of Output Transformations.}\label{paragraph_impact_output_transformation}
To analyze the influence of different commonly used output transformations on \ours, we evaluate \ours in combination with ILM \cite{chen2023understanding} and FM \cite{tsao2024autovp}, referred to as \ours w. ILM and \ours w. FM, respectively. The experimental results obtained using ImageNet-21K pre-trained ViT-B/32 evaluated on CIFAR100 are summarized in Table \ref{table_lm_impact}, from which we make the following observations: \ding{182} \ours achieves the highest performance when using LP as the output transformation across both datasets. \ding{183} Even with identical output transformations used by the baseline methods, \ours consistently outperforms these methods, highlighting the advantage of our dynamic VP adaptation.

\begin{table}[t]
    \centering
        \caption{\textbf{The Impact of Output Transformation.} Performance comparison of \ours using ILM, FM, and LP as output transformations and baseline methods. \ours achieves the best performance across all settings.}
    \label{table_lm_impact}
    \renewcommand{\arraystretch}{1}
    \resizebox{0.9\textwidth}{!}{
    \begin{tabular}{l|c|cc|c}
    \toprule
    \textbf{~ ~ Method} & \textbf{Output Transformatin} & \textbf{Tiny-ImageNet} & \textbf{CIFAR100} & \textbf{Average} \\ 
    \midrule
    LP \textcolor{gray}{[CVPR22]} & LP & 83.88 & 86.01 & 84.95 \\ 
    ILM-VP\textcolor{gray}{[CVPR23]} & ILM & 32.38 & 40.02 & 36.20 \\
    AutoVP\textcolor{gray}{[ICLR24]}  & FM & 82.39 & 85.86 & 84.13 \\
    DAM-VP \textcolor{gray}{[CVPR23]} & LP & 84.90 & 88.10 & 86.50   \\
    SMM \textcolor{gray}{[ICML24]} & ILM & 79.72 & 82.27 & 81.00   \\
    \cellcolor{gray!20} \ours w. ILM  & \cellcolor{gray!20}ILM & \cellcolor{gray!20} 85.57 & \cellcolor{gray!20} 82.44 & \cellcolor{gray!20} 84.01 \\
    \cellcolor{gray!20} \ours w. FM & \cellcolor{gray!20}FM & \cellcolor{gray!20} 86.61 & \cellcolor{gray!20} 89.16 & \cellcolor{gray!20} 87.89  \\
    \cellcolor{gray!20} \ours (Ours) & \cellcolor{gray!20}LP & \cellcolor{gray!20} \textbf{86.65} & \cellcolor{gray!20} \textbf{89.44} & \cellcolor{gray!20} \textbf{88.05}  \\
    \bottomrule
    \end{tabular}
    }
\end{table}

\begin{figure}[t]
  \centering
  \begin{minipage}[b]{0.54\textwidth}
    \centering
    \includegraphics[width=\linewidth]{figures/results/energy_based_compare_resnet18_cifar100_10.pdf}
    \caption{Average predicted probabilities for correctly classified images (left) and incorrectly classified images (right) on CIFAR100 using ImageNet-1K pre-trained ResNet-18. Applying the energy-based dynamic VP adaptation significantly enhances discriminative capability.}
    \label{Figure_energy_based}
  \end{minipage}%
  \hfill
  \begin{minipage}[b]{0.43\textwidth}
    \centering
    \includegraphics[width=\linewidth]{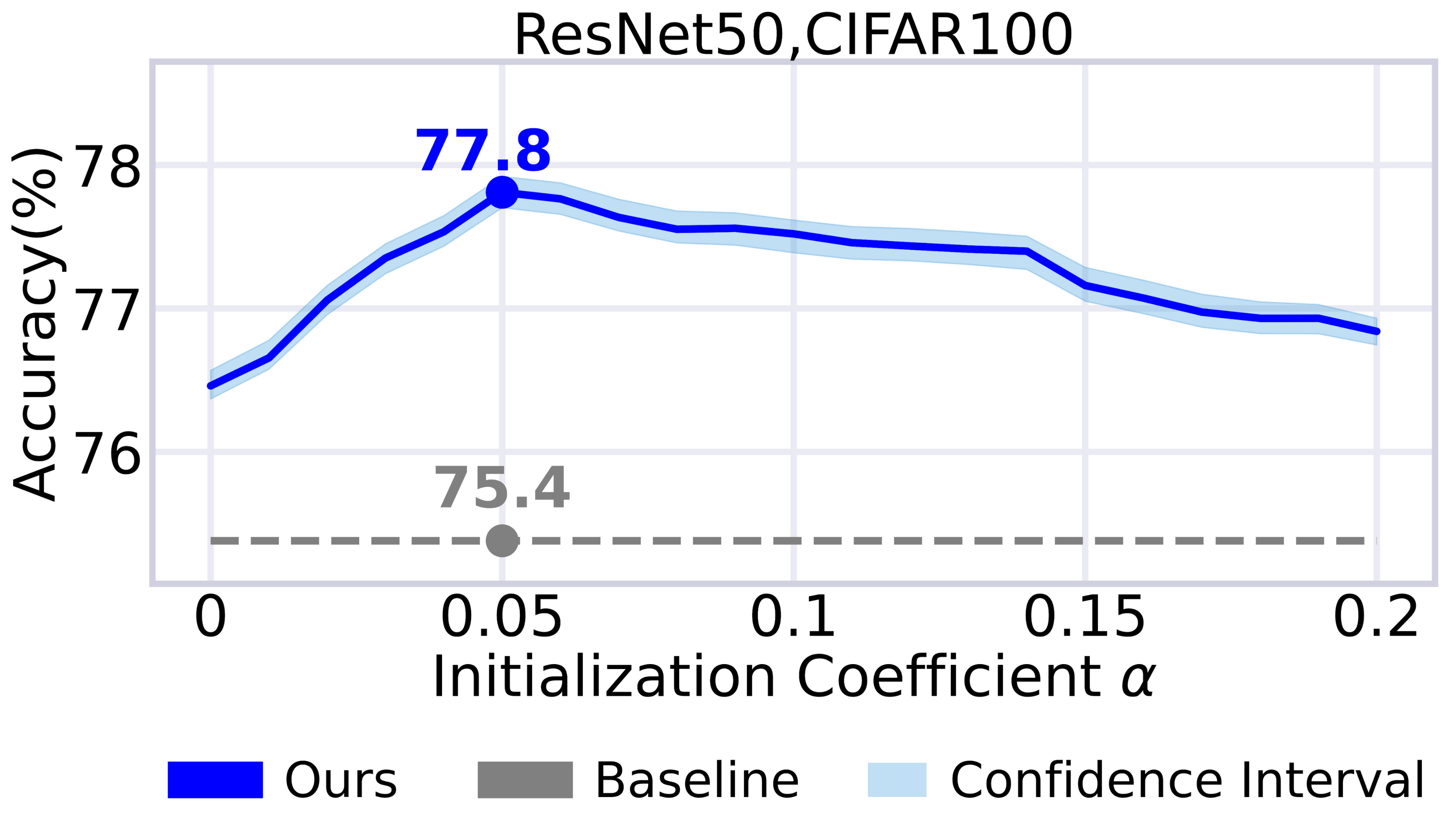}
    \caption{Effect of the initial value of the strength coefficient $\alpha$ on accuracy. The dashed line represents the performance of \ours without dynamic VP adaptation. An $\alpha$ of 0.05 achieves the best performance.}
    \label{figture_resnet50_alpha_acc}
  \end{minipage}
\end{figure}

\paragraph{Impact of the Energy-based Method on VP's Discriminative Power.}
To better understand how the energy-based dynamic VP adaptation in \ours affects discriminative performance, we analyze the predicted probabilities assigned to target labels. Specifically, we conduct experiments using ImageNet-1K pre-trained ResNet-18 evaluated on CIFAR100. After each training epoch, we compare the predicted probabilities obtained with and without the energy-based dynamic adaptation. The results are shown in Figure \ref{Figure_energy_based}. We observe that employing energy-based dynamic adaptation notably increases the average predicted probability for correctly classified images and decreases it for incorrectly classified images. This clearly demonstrates that the energy-based adaptation enhances the discriminative capability of visual prompts in \ours.

\paragraph{Impact of the Strength Coefficient $\alpha$.}
We investigate how the initialization value of the strength coefficient $\alpha$ affects performance using ImageNet-1K pre-trained ResNet-50 evaluated on CIFAR100. Results presented in Figure \ref{figture_resnet50_alpha_acc} indicate that the optimal initialization for $\alpha$ is approximately $0.05$ for the highest accuracy.

\paragraph{Efficiency of \ours.} 
To assess the efficiency of \ours relative to the baseline methods, we evaluate its performance in terms of training epochs, training time, number of VP parameters, inference latency, and inference peak memory. Experiments are conducted using ImageNet-1K pre-trained ResNet-18 and ResNet-50 on the DTD dataset. The results presented in Table \ref{Table_efficiency} highlight several key findings: \ding{182} \ours achieves superior performance while requiring the fewest training epochs, the shortest training time, and the fewest VP parameters. Notably, on ResNet-18, \ours surpasses the best-performing baseline, DAM-VP, by $5.3\%$ while using 1030$\times$ fewer VP parameters. \ding{183} Although \ours exhibits inference latency and memory usage comparable to DAM-VP, its accuracy significantly exceeds that of all baselines. Importantly, unlike existing methods, \ours avoids the auxiliary networks required by SMM, as well as the extensive hyperparameter tuning and complex VP designs of DAM-VP, thereby enhancing practical scalability and deployment simplicity. Given the sizable accuracy gains, substantially faster training, and ease of deployment, the comparable inference cost to DAM-VP is well justified.

\begin{table}[t]
    \centering
        \caption{\textbf{Efficiency Comparison.} Comparison of the training and inference efficiency of \ours compared to the baselines using ResNet-18/ResNet-50 on DTD.}
    \label{Table_efficiency}
    \resizebox{1\textwidth}{!}{
    \begin{tabular}{lll|ccccc|c}
    \toprule
    Method & Type & Model & Epochs & Time & \#VP Params  & Latency & Peak Memory & Acc. \\ 
    \midrule
    ILM-VP\textcolor{gray}{[CVPR23]} & Single-VP & ResNet-18 &  200 & 1.2h & 147K & 1.1ms & 1.4GB & 35.2  \\
    ILM-VP\textcolor{gray}{[CVPR23]} & Single-VP & ResNet-50 &  200 & 1.8h & 147K & 2.5ms & 2.2GB & 40.5  \\
    AutoVP\textcolor{gray}{[ICLR24]} & Single-VP & ResNet-18 & 100 & 0.5h & 101K & 1.1ms & 1.4GB & 54.7 \\
    AutoVP\textcolor{gray}{[ICLR24]} & Single-VP & ResNet-50 & 100 & 0.8h & 101K & 2.6ms & 2.2GB & 65.5 \\
    \midrule
    SMM\textcolor{gray}{[ICML24]}    & Diverse-VP & ResNet-18 &  200  &  1.2h   & 177K &  1.5ms & 2.5GB  &  33.6  \\
    SMM\textcolor{gray}{[ICML24]}  & Diverse-VP & ResNet-50 &  200  &  1.9h   & 177K &  3.3ms & 3.5GB &  41.5  \\
    DAM-VP\textcolor{gray}{[CVPR23]} & Diverse-VP & ResNet-18 &  50  &  1.3h   & 5.2M &  1.7ms & 3.0GB &  61.3  \\
    DAM-VP\textcolor{gray}{[CVPR23]} & Diverse-VP & ResNet-50 &  50  &  2.1h   & 5.2M &  3.7ms & 3.9GB &  69.0  \\
    \cellcolor{gray!20} \ours & \cellcolor{gray!20}Diverse-VP & \cellcolor{gray!20}ResNet-18 & \cellcolor{gray!20} \textbf{20} & \cellcolor{gray!20} \textbf{0.2h} & \cellcolor{gray!20} \textbf{5K}  & \cellcolor{gray!20} \textbf{1.7ms} & \cellcolor{gray!20} \textbf{2.9GB} & \cellcolor{gray!20} \textbf{66.6}\\
    \cellcolor{gray!20} \ours & \cellcolor{gray!20}Diverse-VP & \cellcolor{gray!20}ResNet-50 & \cellcolor{gray!20} \textbf{20} & \cellcolor{gray!20} \textbf{0.3h} & \cellcolor{gray!20} \textbf{5K}  & \cellcolor{gray!20} \textbf{3.7ms} & \cellcolor{gray!20} \textbf{3.7GB} & \cellcolor{gray!20} \textbf{71.1}\\
    \bottomrule
    \end{tabular}}
\end{table}

\vspace{-1mm}
\section{Conclusion}\label{section_conclusion}
\vspace{-1mm}
This paper introduces an efficient and effective adaptive energy-shaped VP method named \ours that utilizes low-rank prompt initialization with energy-based dynamic VP adaptation to generate image-specific VPs. Extensive experiments consistently demonstrate the superior generalization performance and parameter efficiency of \ours compared to SOTA Single-VP and Diverse-VP approaches. Additional investigations provide valuable insights into the effectiveness and training efficiency of energy-based dynamic VP adaptation. Importantly, \ours avoids auxiliary networks, extensive hyperparameter tuning, and multiple VP designs in existing methods, improving scalability in practice and simplicity for deployment.

\clearpage  


%
%
\bibliographystyle{splncs04}
\bibliography{main}
\end{document}